\def\year{2024}\relax
\documentclass[letterpaper]{article} 
\usepackage{aaai22}  

\usepackage{times}  
\usepackage{helvet}  
\usepackage{courier}  
\usepackage[hyphens]{url}  
\usepackage{graphicx} 
\usepackage{natbib}  
\usepackage{caption} 
\DeclareCaptionStyle{ruled}{labelfont=normalfont,labelsep=colon,strut=off} 
\usepackage{algorithm}
\usepackage{algorithmic}
\usepackage{xcolor}

\usepackage{newfloat}
\usepackage{listings}
\floatstyle{ruled}
\newfloat{listing}{tb}{lst}{}
\floatname{listing}{Listing}

\usepackage{booktabs}
\usepackage{multirow}
\title{Hostility Detection in UK Politics: A Dataset on Online Abuse Targeting MPs}

\author {
    Mugdha Pandya,
    Mali Jin, 
    Kalina Bontcheva,
    Diana Maynard    
}
\affiliations {
    Department of Computer Science, University of Sheffield\\
    \{mugdha.pandya, m.jin, k.bontcheva, d.maynard\}@sheffield.ac.uk
}
\usepackage{bibentry}

\begin{document}

\maketitle

\begin{abstract}
Numerous politicians use social media platforms, particularly X, to engage with their constituents. This interaction allows constituents to pose questions and offer feedback but also exposes politicians to a barrage of hostile responses, especially given the anonymity afforded by social media. They are typically targeted in relation to their governmental role, but the comments also tend to attack their personal identity. 
This can discredit politicians and reduce public trust in the government. It can also incite anger and disrespect, leading to offline harm and violence.
While numerous models exist for detecting hostility in general, they lack the specificity required for political contexts. Furthermore, addressing hostility towards politicians demands tailored approaches due to the distinct language and issues inherent to each country (e.g., Brexit for the UK).
To bridge this gap, we construct a dataset of 3,320 English tweets spanning a two-year period manually annotated for hostility towards UK MPs. Our dataset also captures the targeted identity characteristics (race, gender, religion, none) in hostile tweets. We perform linguistic and topical analyses to delve into the unique content of the UK political data. 
Finally, we evaluate the performance of pre-trained language models and large language models on binary hostility detection and multi-class targeted identity type classification tasks. Our study offers valuable data and insights for future research on the prevalence and nature of politics-related hostility specific to the UK.
\end{abstract}

\section{Introduction}
With the rise of social media use among politicians, especially on X, there has been an increase in direct interaction with the public \cite{agarwal2019tweeting}. This interaction, while beneficial for communication and feedback, also exposes politicians to a significant amount of hostile replies due to the anonymity of online platforms \cite{solovev2022hate}. Such hostility is considered a major concern as it erodes public trust in political processes and institutions, which disrupts constructive communication \cite{gross2023online}. Furthermore, it affects the personal lives and mental health of politicians, with online abuse sometimes leading to real-world threats and violence \cite{enock2023tracking}. In extreme cases, sustained hostility has driven some politicians to step down from their roles and retreat from public life altogether \cite{scott2019women}.

Hostility towards politicians is a global phenomenon characterised by widespread misogyny, sexism, and racism. Political and social scientists investigate it through surveys, interviews, and extensive quantitative and qualitative analyses of social media data \cite{haakansson2024explaining,collignon2021increasing,scott2019women}. Their findings indicate that all politicians receive hostility to some degree, but those from minority groups (e.g., Black, female, LGBTQ+) often face increased hostility based on their identity characteristics \cite{hua2020characterizing,carson2024online}.

In Natural Language Processing (NLP), sentiment analysis tools have been used to identify negative tweets and facilitate studies on abuse trends \cite{hua2020characterizing,ward2020turds}.  Although general hostility detection is prevalent, identifying political hostility requires specialised approaches
because political discussions often reflect a country's unique linguistic and cultural characteristics, incorporating regional colloquialism, profanity and prejudices. For example, hostility towards people of colour is more prevalent in the US \cite{lavalley2022occupation}, while the phenomenon of 
Islamophobia is more severe in India \cite{amarasingam2022fight}. Furthermore, hostile posts are frequently tied to popular issues at that time. For instance, recent hostility towards politicians has largely centred around 
illegal immigration in the UK \cite{goodman2024supporting}.

As the body of work on hate speech, abuse and hostility detection in NLP grows \cite{jahan2023systematic}, there has been a move towards developing resources specifically for political hate speech detection across different countries \cite{grimminger2021hate,guellil2020detecting, jafri2023uncovering}. In the UK, Members of Parliament (MPs) represent a wide range of backgrounds, and this diversity is mirrored in the nature of the abusive comments they receive \cite{gorrell2020politicians}. 
Studies have compiled datasets to analyse abuse trends specific to UK politics \cite{southern2021twitter,bakir2024abuse,gorrell2018twits}. As discussed in the related work section, most of these datasets lack hostility-related labels. Among them, only two datasets include labels suitable for automatic political hostility detection, but they do not take into account identity characteristics \cite{agarwal2021hate, ward2020turds}. A third dataset was created in the context of UK political hostility and used to train a classifier, but with a focus on detecting Islamophobia alone \cite{vidgen2020detecting}.

In this paper, we aim to bridge this gap by constructing a high-quality hostility dataset spanning a two-year period to cover diverse political topics in the UK. Our main contributions are:

\begin{itemize}
    \item A publicly available dataset specific to political hostility towards UK MPs consisting of 3,320 tweets with expert annotations for hostility and the targeted identity characteristics (race, gender, religion, none), including individual annotations with confidence scores and gold labels;\footnote{Dataset is available at \url{https://zenodo.org/records/10809695}}

    \item In-depth linguistic and topical analyses to identify linguistic patterns and trending topics in our hostility dataset towards UK MPs;

    \item Evaluation of pretrained language models (PLMs) and large language models (LLMs) on the task of binary hostility identification and multi-class targeted identity type classification in flat and 2-level hierarchical classification settings.
\end{itemize}

Our work distinguishes itself from others by creating a dataset specifically designed for training models to automatically detect political hostility towards UK politicians based on targeted identity characteristics. Through topic analysis, we demonstrate that political hostility is closely tied to contemporaneous events. While this is unsurprising, it nevertheless has important ramifications for training models \cite{jin2023examining}. Linguistic analysis and dataset statistics reveal that the governing political party receives the most hostility proportionally, with race-based hostility being the most prominent among the identity characteristics studied. The extended two-year data collection period of our dataset thus provides a broader range of topics than existing datasets, improving both topic diversity and the generalisability of classifiers trained on this data \cite{jin2023examining}. Additionally, our dataset includes labels for identity characteristics and their combinations, as intersectional abuse is a significant and particularly damaging feature of online hostility \cite{kuperberg2018intersectional,kuperberg2021incongruous}. 

\section{Related Work}

\subsection{Online Hostility}
The rise in social media usage has resulted in a growing amount of hostility \cite{walther2022social,macavaney2019hate}. Consequently, this has prompted NLP research into different hostility detection tasks such as \cite{mansur2023twitter, jahan2023systematic} hate speech, abuse, toxicity, offence, trolling, cyberbullying, etc \cite{pavlopoulos2020toxicity,mathew2021hatexplain}. Hate speech and abuse datasets have labels for hate, abuse, and offence with additional labels for the targeted groups like gender and race \cite{zampieri2019semeval,basile2019semeval}. Toxicity and cyberbullying datasets have label such as harassment, aggression, toxic, etc. \cite{rosa2019automatic,hartvigsen2022toxigen}
However, the definitions of these tasks are very similar \cite{fortuna2020toxic,basile2019semeval}, making it challenging for annotation and for comparison of datasets \cite{zampieri2019semeval, waseem2017understanding}
To avoid this problem, we combine the definitions of these terms in NLP literature into one umbrella term - hostile. While lots of work has been done to combat general hostility detection issues on different social media like Gab, Reddit, X (formerly Twitter), etc, \cite{jahan2023systematic,mollas2022ethos,rieger2021assessing} the problem of the specificity of political hostility data (i.e., language, topic, country) needs more specialised research. 

\subsection{Online Hostility towards Politicians}
Existing work in political hostility typically focuses on qualitative insights or analysis of summary statistics. This work is widespread, and there seems to be an overarching theme of sexism, racism and religious hostility. For instance, female politicians face negative sentiments and attitudes in Japan \cite{fuchs2021normalizing}; in the US, people of colour from the Democratic party and female politicians receive disproportionate hate \cite{solovev2022hate,grimminger2021hate, hua2020characterizing}; MPs in the UK face substantial racial and gender-based abuse \cite{bakir2024abuse, kuperberg2018intersectional}. 

Datasets and machine learning models to detect political hate speech have been created for different countries and their corresponding languages (Arabic in Algeria \cite{guellil2020detecting}, Chinese in Taiwan \cite{wang2022political}, Hindi in India \cite{jafri2023uncovering}). While these datasets can be used to detect country-specific political hate speech, they do not take identity characteristics into account despite their prominence in political hate speech.

\subsection{UK-Specific Hostility towards MPs}
In the UK, political hostility has been studied based on topics and identity characteristics. \citet{bakir2024abuse} and \citet{farrell2021mp} found that abuse towards MPs was at an all-time high during the first year of the COVID-19 pandemic and that women MPs, especially from non-white backgrounds, received higher abuse. 
\citet{gorrell2019race} investigated trends in racial and religious abuse towards MPs around Brexit and also abuse trends leading up to the 2015, 2017 \cite{gorrell2018twits} and 2019 \cite{gorrell2020politicians} General Elections. They found that prominence, Parliamentary events, and MP identity characteristics correlated with receiving abuse. 

A large body of work has focused on gender-based hostility showing that the hostility female MPs face is often in the form of othering, belittling, discrediting, and stereotyping. For example, female electoral candidates' lower success rates were correlated with gender-based harassment \cite{collignon2021increasing}; gender stereotypes and misogyny are reinforced on YouTube through hateful videos and comments \cite{esposito2021dare}; female MPs face more incivility, including stereotyping and questioning credibility than male MPs \cite{southern2021twitter}. However, gender alone is not the only dimension where MPs face hostility. Rather, it is intertwined with other identity characteristics such as age, class, race, and religious beliefs. \cite{kuperberg2021incongruous,esposito2022gender}. 



\subsection{Existing Datasets for UK Political Hostility}
Despite so much awareness about political hostility in the UK, only a small amount of work has been carried out in developing NLP datasets and models specifically for its automatic detection. To the best of our knowledge, there are only 3 datasets that are suitable for this task. Details of these are in Table \ref{tab:datasets}. \citet{agarwal2021hate} created a dataset of 2.5 million tweets collected over 2 months. They used 18 existing social media hate speech classifiers to generate binary hate labels and then studied the topic and MP characteristic trends in political hate speech. However, these hate speech classifiers were not trained specifically on political hate speech data. \citet{vidgen2020detecting} created a dataset and classifier to detect Islamophobia in a political hate speech context. The dataset consists of 4000 tweets collected over 1.5 years with expert manual annotations. While this study uses political social media data, its focus is on Islamophobia alone. Therefore, the labels pertain specifically to Islamophobia rather than to hostility.
\citet{ward2020turds} used sentiment analysis to collect negative tweets from which they created a dataset of 3000 tweets collected over 2.5 months manually annotated for hate and abuse. They used this data to study abuse trends and found that other than identity characteristics, one of the main causes of abuse was reacting to political topics and issues.  

The work presented in this paper differs from existing datasets as it is specifically designed to facilitate the automatic detection of political hostility in the UK, focusing on multiple identity characteristics. Unlike the limited suitable existing datasets, our data collection spans two years, covering a broad range of topics over an extended period. This timeframe is crucial for creating classifiers that generalise more effectively \cite{jin2023examining}. Additionally, we use the dataset to show some preliminary findings about the nature of this hostility, as well as methods to best identify it.

\begin{table*}[t!]
    \centering
    
    \small
    \begin{tabular}{|c|c|c|c|}
    \toprule
         \bf{Dataset} & \bf{Time} & \bf{Tweets} & \bf{Labels} \\\midrule
         Agarwal et al. (2021) & 1 Oct 2017 - 29 Nov 2017 & 2.5 M & hate, not hate\\\bottomrule
         Vidgen et al. (2020) & Jan 2017 - June 2018 & 4000 & none, weak islamophobia, strong islamophobia\\\bottomrule
         Ward et al. (2020) & 14 Nov 2016 - 28 Jan 2017 & 3000 & non-abusive, not-directed, abusive, hate-speech\\\bottomrule
         \textbf{Our dataset} & \textbf{Nov 2020 - Dec 2022} & \textbf{3320} & \textbf{hostile-none, hostile-religion, hostile-gender, hostile-race, non-hostile}\\
    \bottomrule
    \end{tabular}
    \caption{Datasets for automatic UK political hostility detection.}
    \label{tab:datasets}
\end{table*}

\section{Data}
We develop our dataset in 3 steps: data collection, data sampling and annotation.


\subsection{Data Collection}
Following the method used by \citet{bakir2024abuse}, the Twitter (now X) Streaming API is used to follow the accounts of all MPs (568) with active X accounts. We collect four types of tweets relating to each MP between November 2020 and December 2022: the original tweets posted by the MPs, replies to them, retweets of tweets posted by them and retweets created by them. This collection contains over 30 million tweets, which we denote as ${C}$.

\begin{table}[t!]
\centering
    
    \small
    \begin{tabular}{|c|c|c|c|c|}
    \toprule
    \bf{Party}&  \bf{Conservative}&  \bf{Labour}&  \bf{SNP}& \bf{Total}\\
    \midrule    
         Female&  6&  4&  1 & 11 \\ \bottomrule 
         Male& 3 & 4 &  0& 7 \\ \bottomrule
         Non-white& 7 & 4 & 1 & 12\\ \bottomrule 
         White& 2 & 4 & 0 & 6\\ \bottomrule
         Not Christian& 5 &  2&  1& 8\\ \bottomrule 
         Christian&4  & 6 & 0 & 10\\ \bottomrule
    \end{tabular}
    \caption{Statistics of MP identity characteristics and belong parties.}
    \label{tab:dataset}
\end{table}

\subsection{Data Sampling}
The sheer volume of tweets in ${C}$ makes manual annotation infeasible. Therefore, we sample a subset of ${C}$ for the annotation task covering diverse time periods and topics, we denote as ${S}$. We employ the following sampling steps:

\begin{itemize}
    \item We choose a \textbf{subset of 18 MPs} to ensure an annotated dataset with diverse representation of identities and political affiliations. The MPs are selected to ensure the overall pool includes both minority (race: non-White; gender: female; religion: non-Christian) and majority identity groups (race: White; gender: male; religion: Christian).\footnote{The MPs' identity characteristics are based on self-declared public information.} The selected MPs are from the Conservative Party (9 MPs), the Labour Party (8 MPs) and the Scottish National Party (1 MP). Table~\ref{tab:dataset} presents the distribution of identities and parties of these 18 MPs.
    \item A \textbf{long temporal span} was ensured by sampling tweets from the 5 different highest posting activity days for each MP, which occur in ${C}$. 
    \item We exclude duplicated tweets and use an abusive language classifier from \cite{gorrell2020politicians} to identify \textbf{hostility} of all 2.54 million individual tweets. For each of the 5 days, we sample 17 hostile and 20 non-hostile tweets. Therefore, there are potentially 85 hostile and 100 non-hostile tweets per MP which are then manually annotated.
\end{itemize}
In total, ${S}$ contains 3,330 tweets in English. 
\subsection{Data Annotation}
The data annotation process consists of defining the guidelines, performing the annotation task and quality control.

\subsubsection{Annotation Guidelines}
To address the challenge of differentiating between the closely related concepts such as hate, abuse and toxicity, we combined the definitions of these terms from NLP literature into an umbrella term, hostile (see Table~\ref{tab:definitions}). We revised the definitions multiple times to strike a balance between the prescriptive and descriptive paradigm \cite{rottger2021two}. The former enforces definitions and rules that annotators must abide to, while the latter provides guidelines that allow annotators to apply their own understanding. To this end, small focus groups were conducted to discuss the rigidity of the label definitions.

\begin{table*}[t!]
    
    \small
    \begin{tabular}{|l|p{0.4\linewidth}|p{0.4\linewidth}|}
    \toprule
    \bf{Label}&\bf{Definition} &\bf{Example}\\
    \midrule   
    \textbf{Hostile} &  \textbf{Hostility towards a target group or individual. Intended to be derogatory, abusive, threatening, humiliating, inciting violence or hatred towards an individual/members of the group.}&\textit{\textless USER \textgreater and \textless USER \textgreater Put back on your leash were you? There’s a good boy}\\
    \hspace{4mm}Race & \hspace{4mm}Hostility directed at a person/group based on racial background/ethnicity. Including discrimination based on somatic traits (e.g. skin colour), origin, cultural traits, language, nationality, etc..& \hspace{4mm}\textit{\textless USER \textgreater Your in England speak bloody ENGLISH!}\\
    \hspace{4mm}Gender & \hspace{4mm}Hostility directed at a person/group based on their gender. Including negative stereotyping, objectification, using gendered slurs to insult, and threats of a sexual nature. & \hspace{4mm}\textit{\textless USER \textgreater If you can't stand the heat get the hell out of the kitchen next time elect a man to be prime minister, Liz Truss just showed us there are things women can't do.}\\
    \hspace{4mm}Religion & \hspace{4mm}Hostility directed at a person/group based on their religious beliefs. including misrepresenting the truth and criticism of a religious group without a well-founded argument. & \hspace{4mm}\textit{\textless USER \textgreater sick of you tweeting about muslims or any other religion. Your silence speaks the same bullshit, but its ok as Ramadan is over?!?!}\\
    \hspace{4mm}None & \hspace{4mm}Do not refer to gender, race/ethnicity or religion.& \hspace{4mm}\textit{\textless USER \textgreater is the worst human being. I wish someone would shoot her}\\\bottomrule
    \textbf{Not hostile} & \textbf{Posts that are not hostile. a tweet containing profanity is not hostile unless its context makes it so.} & \textit{\textless USER \textgreater will make a bad PM. Please don’t turn this into a race war. Please notice that he is a terrible politician}\\\bottomrule
    \end{tabular}
    \caption{Hostility taxonomy with targeted identity type definitions and examples.}
    \label{tab:definitions}
    \end{table*}


We consider political hostility detection as a hierarchical classification task. Given a tweet $t$, the aim is to classify $t$ based on hostility (binary classification) and the target identity characteristics (multiclass classification). We formulate the task in a hierarchical manner similar to existing datasets like OffensEval \cite{zampieri2019semeval} and HatEval \cite{basile2019semeval}. First, $t$ is classified into two hostility labels: hostile and not hostile. If $t$ is classified as hostile, then it will be further classified into one of the four target identity characteristic labels: religion, gender, race and none. Table~\ref{tab:definitions} shows the definitions of each category and example tweets. Note that hostility can be intersectional (i.e., target multiple identity characteristics simultaneously), so a tweet can have more than one identity label. To provide a measure of reliability of each annotation, we include a confidence score of 1 to 5, from very low confidence to extreme confidence for both hostility and identity characteristic labels. Table~\ref{tab:confidence} in the appendix presents the confidence scores with explanations.

%

\subsubsection{Annotation Method}
The annotation task was conducted in three steps: training, testing, and annotation. Steps 1 and 2 ensured high-quality annotations. The entire annotation process was conducted using the collaborative web-based annotation tool Teamware 2 \cite{wilby2023gate}.
\begin{figure}[t!]
  \centering
  \includegraphics[width=0.8\linewidth,height=0.6\linewidth]{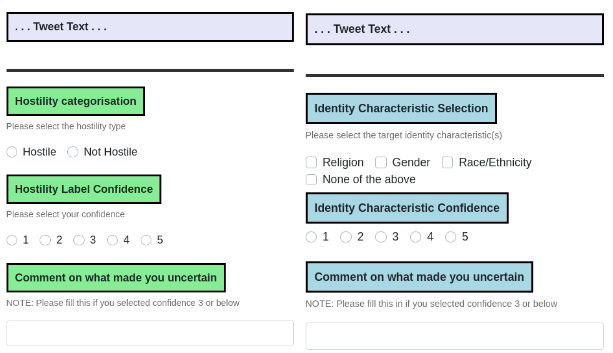}
  \caption{Annotation platform user interface.}
  \label{fig:UI}
\end{figure}
\begin{enumerate}
    \item \textbf{Training sessions}: Training sessions were conducted in which annotators received in-person presentations explaining label definitions with detailed examples. Annotators were also guided on setting up their annotator account and familiarising themselves with the platform.
    \item \textbf{Testing sessions}: Each annotator then underwent a short test to ensure a proper understanding of the task and guidelines, consisting of 20 tweets covering all the labels. Annotators were required to label at least 14 of the 20 annotations correctly. Finally, annotators were provided with the correct answers as well as explanations. 
    \item \textbf{Annotation}: Once annotators passed the test, they were added to the actual annotation task and were shown tweets sequentially. Figure~\ref{fig:UI} shows the platform user interface. 
\end{enumerate}

\subsubsection{Annotation Task Quality}
A number of steps were taken to ensure high-quality manual annotations. Annotators were recruited from postgraduate study courses in Politics and Computer Science. The only prerequisite was that they had to be familiar with UK politics and colloquialisms. We placed no restriction on age, gender, ethnicity, etc. so as to not bias the labels. We contacted potential annotators by emailing the respective course groups. Each annotator was paid 30 GBP for the annotation of 200 tweets. We recruited a total of 48 annotators. Each tweet in ${S}$ is labelled by 3 annotators. 

During the task, annotators were instructed to look up unfamiliar terms and slang. Each annotator was allowed to annotate only 200 tweets in total, and the task did not need to be completed in one sitting. This allowed annotators to take breaks and prevented them from getting overly desensitised to the hostile content. 

A manual analysis of the annotation results established that some annotators had incorrectly confused the race and religion labels in cases where Muslims and Jews were being targeted. Therefore, expert annotators corrected this small number of annotations.

\subsection{Dataset}
The fully annotated dataset consists of 3,320 tweets in total after removing posts that contain URLs or user mentions only. We use 3 sets of gold labels for our modelling experiments, as follows:
\begin{itemize}
    \item \textbf{Set 1:} The gold labels were assigned based on the majority vote, i.e. the label which has at least 2 annotations out of 3. For the cases where multiple identity labels were chosen (intersectional), an expert manually assigned the dominant label in their opinion.     
    \item \textbf{Set 2:} Annotations with confidence \textless 3 were removed, and gold labels are derived from the remaining annotations. In the cases where only one annotation remained for a tweet, that was selected as the gold label. When there were 2 annotations, the annotation with the higher confidence was selected. If both had equal confidence, an expert manually assigned the dominant label in their opinion. If all 3 annotations remained, we used the majority vote method as used in Set 1.
    \item \textbf{Set 3:} To investigate intersectionality in the data and model performance,
we used the same method as Set 2 for high-level hostility labels. For the lower-level identity labels, if there was an intersectional label with confidence \textgreater 2, we chose that as the gold label. We had no cases of different intersectional labels with confidence \textgreater 2. 
\end{itemize}
\begin{table}[t!]
    \small
    \centering
    \begin{tabular}{|l|l|c|c|c|}
    \toprule
    \bf{Hostility}&\bf{Identity} & \bf{Set 1} & \bf{Set 2} & \bf{Set 3}\\
    \midrule
    \multirow{ 5}{*}{Hostile}&Religion & 36 & 41 &52 (26)\\
    &Gender & 108 & 119 & 119 (22)\\
    &Race & 188 & 182&205 (38)\\
    &None & 1135 & 1112&1121 (0)\\
    &Total & 1467 & 1454 & 1454 (43)\\\bottomrule
    Not Hostile & Total & 1853 & 1866 & 1866\\\bottomrule
    \bottomrule
    \multirow{ 2}{*}{Fleiss' $\kappa$} & Hostility &0.68 &0.79 & 0.79\\
    & Identity & 0.51 & 0.65 & 0.47\\\bottomrule
    \end{tabular}
    \caption{Label counts for each set. For Set 3, the value in parentheses shows the count of identity-based hostility that comes from intersectional labels.}
    \label{tab:hostile}    
\end{table}

%
    

Table~\ref{tab:hostile} shows the statistics of each set. The top 6 rows present the frequency of each label for each set. In general, non-hostile tweets account for the largest proportion, followed by no identity and race-based hostile tweets. Set 3 includes intersectional labels, but there are only 43 of these, of which 5 target both religion and gender, 21 target both religion and race, and 17 target both gender and race. The bottom 2 rows present the Fleiss' $\kappa$ annotator agreement score \cite{fleiss1971measuring} for hostility and target identity annotation. Set 2 exhibits the highest $\kappa$-value for both hostility (0.79) and identity (0.65) annotation tasks, which belong to substantial agreement \cite{artstein2008inter}. This suggests selecting annotations based on confidence scores helps to improve the quality of the dataset. 

\begin{figure}
    \centering
    \includegraphics[width=0.6\linewidth,height=0.4\linewidth]{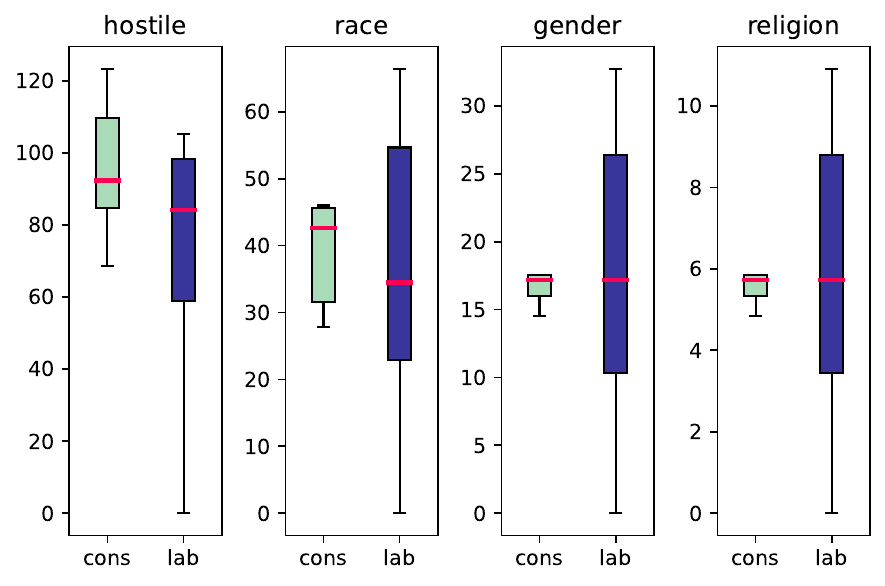}
    \caption{Comparing political party-based differences in the amount and type of hostility received}
    \label{fig:stats_party}
\end{figure}

\begin{figure}
    \centering
    \includegraphics[width=0.7\linewidth,height=0.6\linewidth]{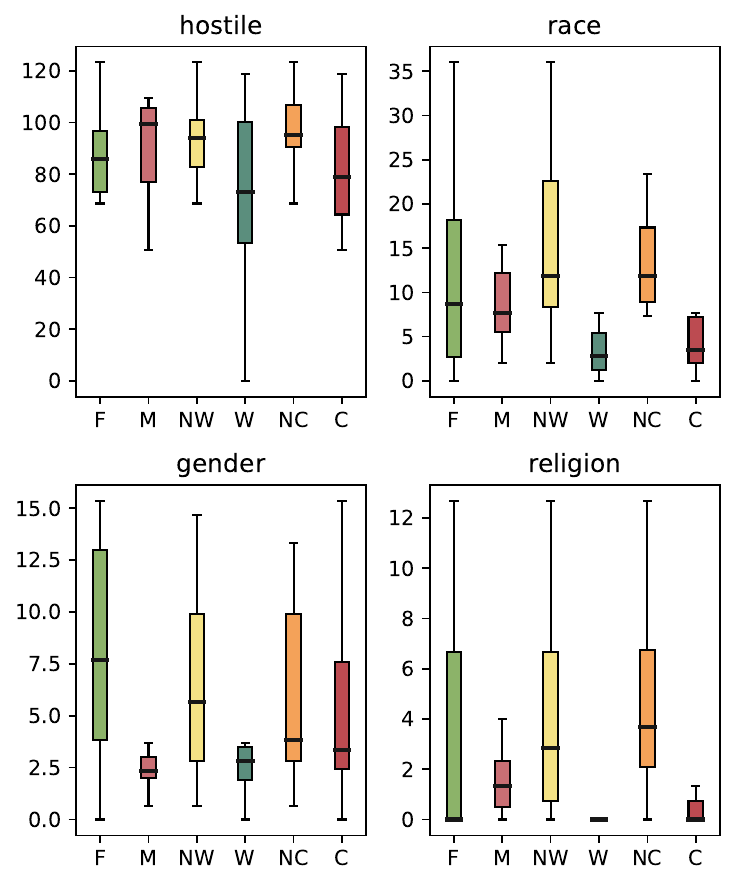}
    \caption{Comparing identity-based differences in the amount and type of hostility received}
    \label{fig:stats}
\end{figure}

Figure~\ref{fig:stats_party} and Figure~\ref{fig:stats} show the 
amount and type of hostile tweets MPs receive based on their political party and identity group. The horizontal pink (Figure~\ref{fig:stats_party}) and black (Figure~\ref{fig:stats}) lines mark the mean value for each group. 
On average, MPs belonging to the Conservative Party receive more race-based hostility.
For gender and religion-based hostility, on average, MPs from both parties receive a similar amount of hostility. However, there are some MPs from the Labour Party who receive more identity-based hostility than others (e.g. Diane Abbott, David Lammy, etc.). Due to only one MP in our study belonging to SNP, we do not include SNP in this comparison.  

In Figure~\ref{fig:stats}, as expected, we see that while male (M) MPs receive more hostile tweets, 
female (F) MPs are disproportionately subjected to gender-based hostility. 
Similarly, non-white (NW) and non-Christian (NC) MPs face significantly higher levels of general and both race and religion-based hostility. Interestingly, we see that MPs from racial and religious minority groups consistently receive more general hostility and identity-based hostility (consistently higher mean values for all types of hostile tweets), than their white (W) or Christian (C) counterparts. This highlights the issues of intersectional hostility \cite{kwarteng2022misogynoir} wherein different minority groups intersect with and reinforce each other. 

\section{Data Characterisation}
\subsection{Linguistic Analysis}

To investigate the difference between both the use of language and the content of hostile and not hostile tweets, we conduct a comparative linguistic analysis. We use the Bag of Words (BOW) model and Linguistic Inquiry and Word Count (LIWC) Dictionary \cite{boyd2022development} to identify linguistic patterns. Then a univariate Pearson's correlation test is used to identify which of these linguistic patterns significantly correlate with hostile and not hostile tweets respectively. During pre-processing, URLs and user @mentions in the tweets are replaced with special tokens (\textless URL \textgreater and \textless USER \textgreater, respectively), and stop words are removed using NLTK \cite{bird2009natural}.

\begin{figure}[t!]
  \centering
  \includegraphics[width=0.7\linewidth]{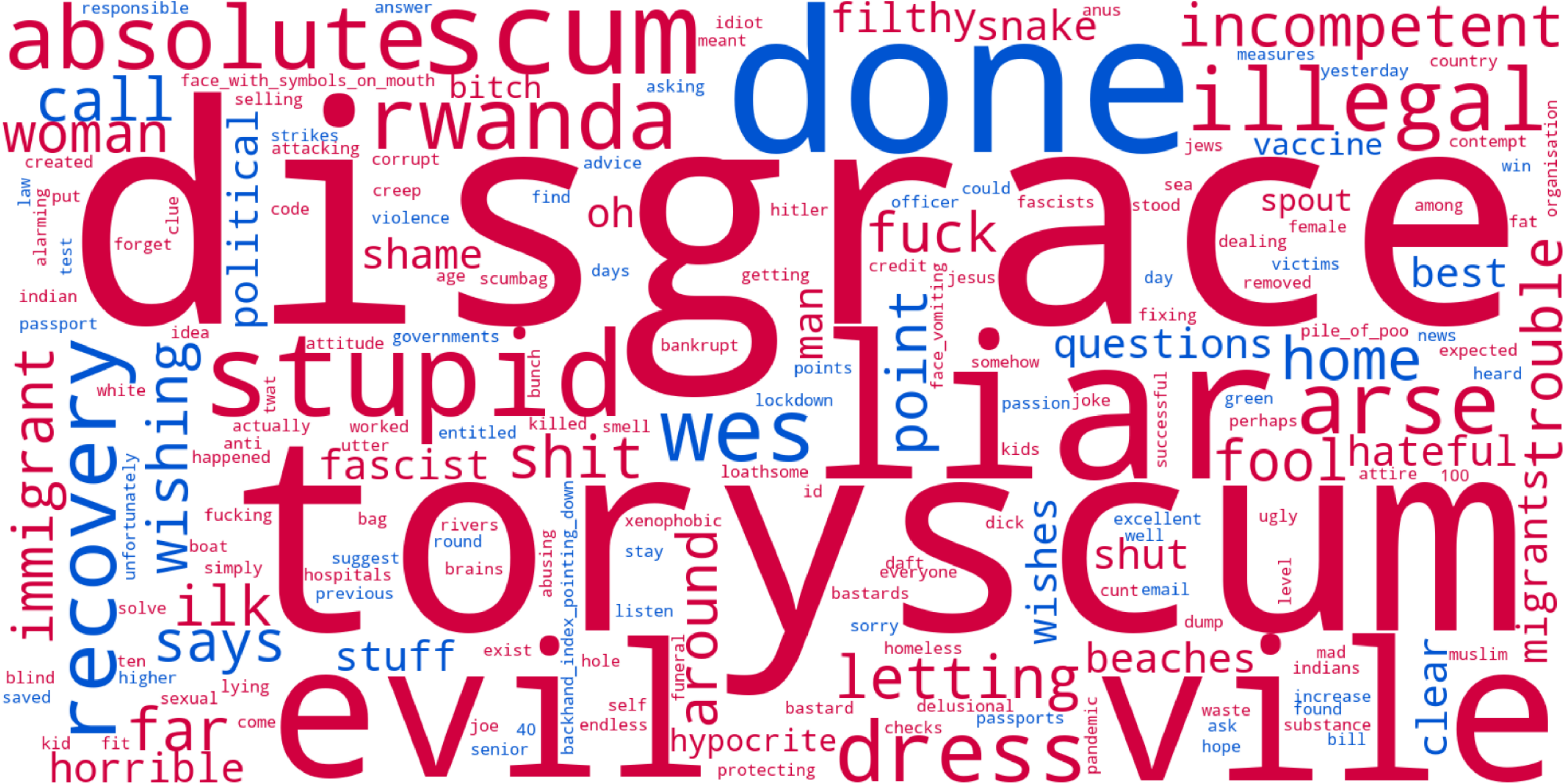}
  \caption{Top 100 BOW unigrams associated with \textcolor{red}{hostile} and \textcolor{blue}{non-hostile} tweets. The larger the text size, the higher the Pearson correlation coefficient $r$, and vice versa.}
  \label{fig:bow_uni}
\end{figure}

\begin{figure}[t!]
  \centering
  \includegraphics[width=0.7\linewidth]{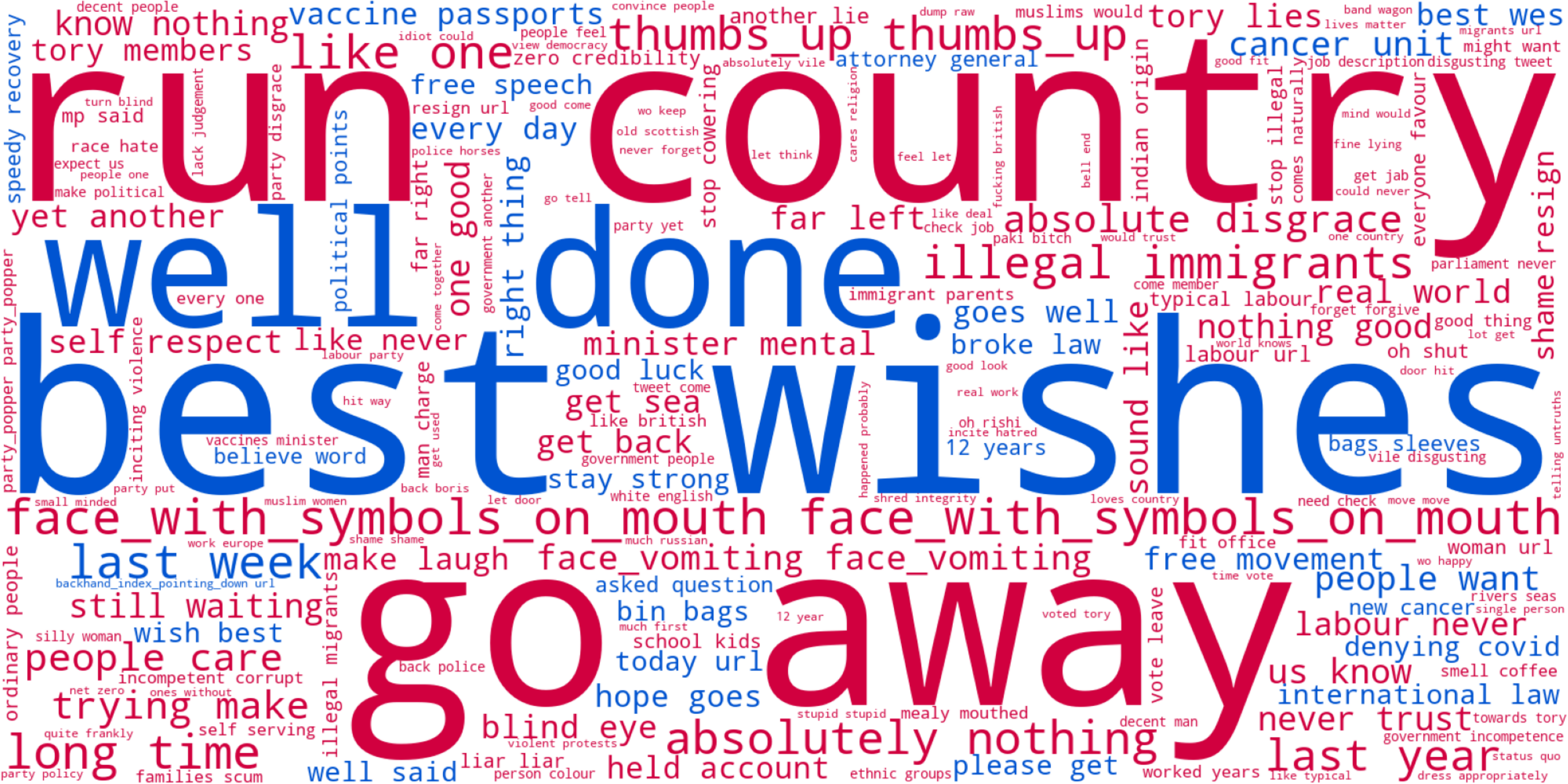}
  \caption{Top 100 BOW bigrams associated with \textcolor{red}{hostile} and \textcolor{blue}{non-hostile} tweets. The larger the text size, the higher the Pearson correlation coefficient $r$, and vice versa.}
  \label{fig:bow_bi}
\end{figure}

\subsubsection{BOW}
We begin by employing the Bag-of-Words (BOW) model to represent each post as a TF-IDF weighted distribution over a vocabulary of the 3,000 most frequent unigrams and bigrams. To visually highlight the differences in BOW features associated with hostile and not hostile tweets, we create word clouds (see Figure~\ref{fig:bow_uni} and Figure~\ref{fig:bow_bi} for unigrams and bigrams, respectively). 

First of all, we observe that hostile tweets are characterised by negative and abusive words and phrases such as ``scum'', ``vile'', ``incompetent'', ``evil'', ``stupid'', ``nothing good'' and ``absolute disgrace''. These tweets are mostly posted to vent anger or dissatisfaction at politicians, ranging from questioning their abilities and distrusting their policies to insulting their personal traits. There are also emojis like ``face\_with\_symbols\_on\_mouth,'' and ``face\_vomiting,'' which represent the use of profanity and disgust. Here is an example 
from our dataset:
\begin{quote}
    Tweet 1: \textit{\textless USER\textgreater Some in the Cabinet are incompetent. Some are corrupt. Some are evil. You are all three. You stand for nothing that is good. I have nothing but contempt and disgust for you.}
\end{quote}
Secondly, phrases such as ``go away'', ``shame resign'', ``run country'' in hostile tweets suggest that much of the hostility is directed at the Conservative Party, which is the ruling party. Here is an example requesting the MP to resign:
\begin{quote}
    Tweet 2: \textit{
Too late with \textless USER\textgreater in charge \& his cabinet of mendacious halfwits. Demand his resignation.}
\end{quote}
Also, phrases such as ``liar'', ``hypocrite'', ``corrupt'', ``never trust'', ``know nothing'' and ``another lie'' indicate a general distrust in the MPs. Furthermore, we notice some trending topics in hostile tweets, such as ``vaccine passports'' and ``illegal immigrants'', which reveal specific issues that cause dissatisfaction. The example tweet expresses the anger at policies relating to illegal immigration:
\begin{quote}
    Tweet 3: \textit{\textless USER \textgreater Just what would you do about the illegal immigration welcome them with open arms just wish we could send you to Rwanda and your filthy son}
\end{quote}




For non-hostile tweets, the correlation $r$ is lower (as can be seen from the text size in Figure~\ref{fig:bow_uni}). However, they are correlated with words and phrases such as ``excellent'', ``recovery'', ``best wishes'' and ``well done''. These suggest that not hostile tweets often contain appreciative and positive emotions towards MPs. Some phrases such as ``asked questions'' and ``free movement'' indicate users' attempts to voice their political concerns. The following tweet is an example conveying appreciation to the MP:
\begin{quote}
    Tweet 4: \textit{\textless USER \textgreater was on fire! Another spectacular debate. Well done sir!}
\end{quote}

\subsubsection{LIWC}
Each tweet is also characterised using psycho-linguistic categories from the LIWC 2022 dictionary \cite{boyd2022development}. Table~\ref{tab:liwc} presents the top 10 LIWC categories most strongly correlated with hostile and not hostile tweets. 

\begin{table}[t!]
    \small
    \centering
    \begin{tabular}{|l|c|l|c|}
    \toprule
    \bf{Hostile} & \bf{\textit{r}} & \bf{Not hostile} & \bf{\textit{r}}\\
    \midrule
    socrefs & 0.186 & Tone & 0.192\\\bottomrule
    you & 0.181 & OtherP & 0.189 \\\bottomrule
    swearwords & 0.162 & AllPunc & 0.183\\\bottomrule
    clout & 0.160 & focuspast & 0.133\\\bottomrule
    tone\_neg & 0.160 & comm &0.104\\\bottomrule
    moral & 1.51 & prosoc & 0.084 \\\bottomrule
    affect & 0.142 & polite & 0.064\\\bottomrule
    ppron & 0.131& i & 0.063\\\bottomrule
    ethnicity & 0.111 & work & 0.062\\\bottomrule
    sex & 0.109 & tone\_pos & 0.061\\
    \bottomrule
    \end{tabular}
    \caption{Top 10 LIWC categories associated with hostile and not hostile tweets sorted by Pearson's correlation ($r$) between the normalised frequency and the labels. All correlations are significant at $p$ \textless .001, two-tailed t-test.}
    \label{tab:liwc}
\end{table}

Similar to BOW, We find that (unsurprisingly) hostile tweets tend to have a negative tone (\textit{tone\_neg}) and convey negative emotions like anger and sadness (\textit{affect}). They contain more assertive and judgemental language (\textit{clout} and \textit{moralisation}). Unsurprisingly, they also contain more swear words (\textit{swearwords}) and sexual terms (\textit{sex}). Interestingly, race-related (\textit{ethnicity}) terms are frequent, suggesting that hostility is often related to race. 
The following example shows race-based hostility towards the MP from the dataset:
\begin{quote}
    Tweet 5: \textit{\textless USER \textgreater What about black violence! Your just another race divider. Marxists like you have ruined this country and divided it further!}
\end{quote}




We notice that non-hostile tweets are highly correlated with the tone marker (\textit{tone}) in general and, specifically, a positive tone (tone\_pos). The tweets are polite, adhere to social norms (\textit{prosoc} and \textit{polite}), and are more communicative (\textit{comm}), often consisting of explanations, feedback and questions. They also express concerns about work, jobs, schooling, etc. (\textit{work}).  Below is an example tweet expressing concerns about the new scheme: 
\begin{quote}
    Tweet 6: \textit{\textless USER \textgreater please consider the effect of this new scheme before putting it into action. We have lost jobs and suffered a lot during covid. The UK economy will not recover. We need to think of our next steps very carefully}
\end{quote}

\begin{table}[t!]
\small
\centering
    \begin{tabular}{|p{0.2\linewidth}|p{0.7\linewidth}|}
    \toprule
         \bf{Topic} & \bf{Representative Words}\\
    \midrule
         Brexit& brexit,  uk, ireland, eu, europe, leave, deal, citizens, free, border\\ \bottomrule 
         Illegal immigration & refugees, illegal, boats, rwanda, immigrants, asylum, raped, terrorists, seekers, migrants \\ \bottomrule 
         Conservative party & tory, conservative, resign, vote, rishi, scum, tories, johnsonout, torysewageparty, cabinet\\ \bottomrule 
         Labour party& labour, starmer, voters, corbyn, party, win, mps, election, abbott, protest\\ \bottomrule 
         COVID-19& covid, virus, vaccine, lockdown, died, pandemic, mask, vulnerable, jab, nhs\\ \bottomrule 
         Cost of living crisis & economy, bills, winter, job, tax, inflation, energy, nhs, heating, gas\\
         \bottomrule
    \end{tabular}   
    \caption{Topic groups and representative words derived from BERTopic}
    \label{tab:topic_censored_nomps}
\end{table}

\subsection{Topic Analysis}
We perform topic analysis using BERTopic \cite{grootendorst2022bertopic} after removing stop words with NLTK \cite{bird2009natural}. Because of the dominance of MP names and profanity, the topics are rather unclear. Table \ref{tab:topic_censored_nomps} shows the top 6 topic groups and their representative words after removing these terms. The topics relate to major events and issues in the UK, such as Brexit (e.g., ``europe'', ``border''), illegal immigration (e.g., ``refugees'', ``terrorists''), and the cost of living crisis (e.g., ``bills'', ``tax'', ``inflation''). The following example is a hostile tweet expressing anger due to increased costs of bills: 
\begin{quote}
    Tweet 7: \textit{\textless USER \textgreater What planet do you live on in your head ? You haven’t saved the day. Fuel is still +40\% on years average. Energy bills are increasing 50\%. We’re all still fucked. Make it make sense !!!}
\end{quote}

Other popular topics are the two main political parties in our dataset (Conservative and Labour). However, the ruling Conservative party is likely to receive more hostility based on the negative terms from representative words such as ``scum'', ``johnsonout''. Here is an example of hostile tweets mentioning the Conservative Party:
\begin{quote}
    Tweet 7: \textit{\textless USER \textgreater \textless USER \textgreater this you? You ludicrous pork Hay-bale. You bin bag full of custard. \#ToryScum \#ToryCriminalsUnfitToGovern it's true. \textless USER \textgreater shouldn't apologise you scum}
\end{quote}



\begin{figure}[t!]
  \centering
  \includegraphics[width=0.8\linewidth]{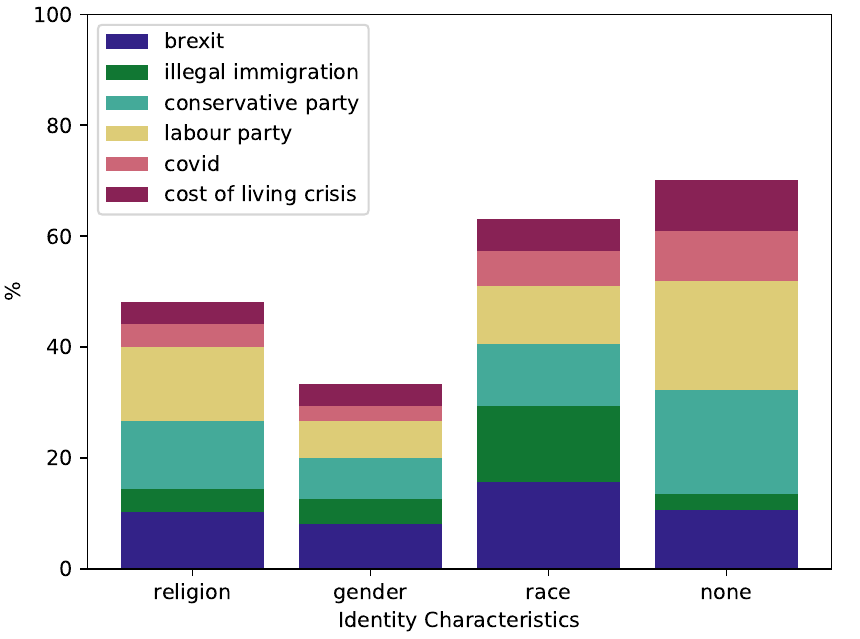}
  \caption{Proportion of topic-related tweets belonging to each identity characteristic label}
  \label{fig:topic_id}
\end{figure}

Most topics appear in the same proportion in hostile tweets as they do in non-hostile tweets. The exception is ``illegal immigration'' which appears twice as much in hostile tweets than in non-hostile tweets. Figure~\ref{fig:topic_id} shows the proportion of topic-related tweets belonging to identity-based hostility. Looking at the distribution of ``illegal immigration'' and ``Brexit'', they appear mainly in race-based hostile tweets. While the ``Conservative party'' and ``Labour party'' topics contribute to race-based hostile tweets, they appear much more frequently in non-race, gender or religion-based hostility.

While all the tweets relate to MPs, they still naturally fall into topics related to current issues at the time. Due to its 2-year span, the dataset thus covers a wide range of topics. This topic characterisation means that the dataset could eventually be used for analysis and comparison of hostility in relation to different issues over time. 
 
\section{Online Hostility Detection}

Given a text snippet, we define online hostility detection as two classification tasks: (1) binary hostility identification (if a tweet contains hostility or not) and (2) multi-class classification for classifying if a tweet contains one of the four identity-based hostility types (i.e., religion, gender, race, none) or no hostility at all. For multi-class classification, we compare two classification methods, namely flat and two-level hierarchical classification.

\paragraph{Flat Classification} Given a tweet, models classify it as religion-based hostility, gender-based hostility, race-based hostility, hostility with no protected identity characteristics (none-hostility) or not hostility. 

\paragraph{Two-level Hierarchical Classification} Given a tweet, the first classifiers identify if it is hostile or not. Then, the second classifiers classify the identity types of the hostile tweets (religion, gender, race or none) based on predictive results from the first classifiers.

\subsection{Predictive Models}
We use three PLMs for binary hostility identification and multi-class classification. Also, we evaluate two widely used LLMs on identifying hostile tweets as well as their targeted identity types.

\paragraph{BERT} We fine-tune Bidirectional Encoder Representations from Transformers (BERT) \cite{devlin2018bert}  by adding a classification layer with softmax activation function on top of the [CLS].

\paragraph{RoBERTa} We fine-tune RoBERTa \cite{liu2019roberta} model in a similar way to BERT.

\paragraph{RoBERTa-Hate} Similarly, we fine-tune a domain adaptation model, RoBERTa-Hate\footnote{\url{https://huggingface.co/cardiffnlp/twitter-roberta-base-hate-latest}}, which is trained on 13 different hate speech datasets in the English language including political content.

\paragraph{LLaMA-3-8B (LLaMA)} We use the instruction tuned LLaMA 3 8B model through the Hugging Face platform\footnote{\url{https://huggingface.co/meta-llama/Meta-Llama-3-8B}}. We provide the model with a sequence of texts and a prompt with a task description to guide its output.

\paragraph{GPT-3.5 (GPT)} Similarly, we use the GPT-3.5 model\footnote{\url{https://platform.openai.com/docs/models/gpt-3-5-turbo}} with an API key providing texts and corresponding prompts.

\subsection{Experimental Set-up}
Tweets are pre-processed where URLs and user @mentions are replaced with special tokens (\textless URL \textgreater and \textless USER \textgreater, respectively). 
For evaluation, we report average Accuracy, Precision, Recall and macro F1 over 5 folds with standard deviations. Details of model parameters can be found in the appendix.

For LLMs, we input the prompt to specify the task for binary hostility identification and multi-class flat classification: (1) \textit{Classify the tweet as hostile or not hostile} and (2) \textit{Classify the tweet into religion-based hostile, gender-based hostile, race-based hostile, other-hostile or not hostile} with (\textbf{LLaMA w/ Def.}, \textbf{GPT w/ Def.}) or without definitions (\textbf{LLaMA}, \textbf{GPT}) of each category (see Table \ref{tab:definitions} for definitions). For 2-level hierarchical classification, we input the prompt based on the outputs from the binary hostility identification: \textit{Classify the tweet as hostility based on race, gender, religion or other}. 
For a fair comparison, we also report the average performance over 5 folds with the same data in each fold.

\begin{table}[t!]
   
    \centering
    \scriptsize
    \begin{tabular}{|l|c|c|c|c|}
    \toprule
    Model & Accuracy & Precision & Recall & F1\\
    \midrule
    \multicolumn{5}{|c|}{Set 1}\\
    \bottomrule
    BERT & 66.96$\pm$1.35 & 66.55$\pm$1.45 & 65.75$\pm$1.32 & 65.84$\pm$1.35\\
    RoBERTa& 68.13$\pm$0.83 & 68.04$\pm$0.62 & 67.55$\pm$0.51 &67.44$\pm$0.48\\
    RoBERTa-Hate & 67.38$\pm$1.51 & 67.47$\pm$1.15 & 67.10$\pm$0.66 & 66.84$\pm$1.09\\
    \bottomrule   
    \multicolumn{5}{|c|}{Set 2}\\
    \bottomrule
    BERT & 72.47$\pm$3.56 & 72.27$\pm$3.82 & 71.62$\pm$3.22 & 71.77$\pm$3.37\\
    RoBERTa & 71.77$\pm$3.37 & 72.26$\pm$2.05 & 69.15$\pm$2.99 & 68.86$\pm$3.65\\
    RoBERTa-Hate & 72.27$\pm$3.82 & \bf{73.44}$\pm$1.00 & \bf{73.16}$\pm$1.44 & \bf{73.03}$\pm$1.27\\
    \bottomrule
    LLaMA & 71.30$\pm$0.96 & 71.17$\pm$0.86 & 71.44$\pm$0.86 & 71.11$\pm$0.91\\
    LLaMA w/ Def. & \bf{73.55}$\pm$1.39 & 73.21$\pm$1.42 & 72.76$\pm$1.43 & 72.91$\pm$1.43\\
    GPT & 60.57$\pm$1.93 & 69.97$\pm$1.21 & 64.20$\pm$1.72 & 58.67$\pm$2.41 \\
    GPT w/ Def. & 70.69$\pm$1.27 & 71.90$\pm$1.29 & 71.85$\pm$1.28 & 70.69$\pm$1.27 \\
    \bottomrule
    \end{tabular}
     \caption{Accuracy, Precision, Recall and macro F1-Score (F1) for binary hostility identification ($\pm$ std. dev.). Best results are in bold.}
    \label{tab:binary_predictive_results}
\end{table}

\subsection{Results}

\paragraph{Binary Hostility Identification} Table \ref{tab:binary_predictive_results} presents the predictive results of all models on binary hostility identification using Set 1 and Set 2. We exclude Set 3 because the intersectional labels in identity type annotation do not affect binary labels (i.e., hostile or not hostile). Overall, RoBERTa-Hate on Set 2 achieves the best performance among all models, reaching a macro F1 score up to 73.03 (in bold). Then, we observe that models trained on Set 2 achieve better performance than those trained on Set 1 (e.g., 68.86 vs. 67.44 for RoBERTa on Set 2 and Set 1), highlighting the importance of selecting annotations based on confidence scores. Also, the domain adaptation model (i.e., RoBERTa-Hate) outperforms the vanilla models on Set 2 (e.g., 68.86 F1 for RoBERTa vs. 73.03 F1 RoBERTa-Hate) and has comparable performance with the vanilla models on Set 1 (e.g., 67.44 F1 for RoBERTa vs. 68.84 F1 for RoBERTa-Hate).\footnote{We also evaluate Set 1 and Set 2 on the same test set with the same labels (we exclude Set 3 since adding intersectional labels leads to different test sets). RoBERTa and RoBERTa-Hate using Set 2 achieve better results than using Set 1 (72.46 vs. 71.11 and 74.10 vs. 73.26 accordingly).}

\begin{table}[t!]
    \centering
    \scriptsize
    \begin{tabular}{|l|c|c|c|c|}
    \toprule
    Model & Accuracy & Precision & Recall & F1\\
    \midrule
    \multicolumn{5}{|c|}{Flat Classification}\\
    \bottomrule
    \multicolumn{5}{|c|}{Set 1}\\
    \bottomrule
    BERT & 61.93$\pm$2.03 & 31.42$\pm$2.14 & 28.31$\pm$2.36 & 28.18$\pm$2.91\\
    RoBERTa & 63.13$\pm$0.35 & 24.40$\pm$0.32 & 26.03$\pm$0.20 & 20.79$\pm$0.21\\
    RoBERTa-Hate & 64.01$\pm$1.41 & 45.81$\pm$6.34 & 34.74$\pm$3.90 & 36.67$\pm$4.54\\
    \bottomrule
    \multicolumn{5}{|c|}{Set 2}\\
    \bottomrule
    BERT & 65.93$\pm$1.68 & 37.12$\pm$1.62 & 34.77$\pm$2.79 & 35.07$\pm$2.72\\
    RoBERTa & 67.26$\pm$1.05 & 33.01$\pm$5.96 & 32.25$\pm$3.95 & 32.04$\pm$4.68\\
    RoBERTa-Hate & \bf{69.64}$\pm$1.24 & 50.82$\pm$7.61 & 41.66$\pm$4.90 & 43.31$\pm$5.65\\
    \bottomrule
    LLaMA & 55.48$\pm$2.29 & 42.01$\pm$5.38 & 46.60$\pm$8.18 & 38.09$\pm$5.69\\
    LLaMA w/ Def. & 61.33$\pm$1.41 & 48.61$\pm$3.58 & 52.39$\pm$3.68 & 47.02$\pm$3.25\\
    GPT & 65.57$\pm$1.77 & 52.64$\pm$4.21 & 55.02$\pm$5.21 & 52.65$\pm$4.45\\
    GPT w/ Def. & 65.03$\pm$1.71 & 53.70$\pm$2.71 & 56.89$\pm$2.69 & 54.74$\pm$2.22 \\
    \bottomrule
    \multicolumn{5}{|c|}{Set 3}\\
    \bottomrule
    BERT & 65.03$\pm$2.31 & 19.45$\pm$3.80 & 18.43$\pm$2.03 & 18.23$\pm$2.66\\
    RoBERTa & 66.66$\pm$1.72 & 20.30$\pm$3.68 & 20.43$\pm$3.44 & 20.08$\pm$3.64 \\
    RoBERTa-Hate & 68.70$\pm$2.64 & 28.55$\pm$8.12 & 23.51$\pm$3.70 & 23.98$\pm$4.49\\
    \bottomrule
    \bottomrule
    \multicolumn{5}{|c|}{Hierarchical Classification}\\
    \bottomrule
    \multicolumn{5}{|c|}{Set 1}\\
    \bottomrule
    BERT & 60.78$\pm$1.00 & 27.44$\pm$5.76 & 25.60$\pm$2.01 & 24.79$\pm$2.14\\
    RoBERTa & 61.99$\pm$1.32 & 37.66$\pm$9.18 & 27.53$\pm$2.08 & 28.87$\pm$2.44\\
    RoBERTa-Hate & 62.47$\pm$2.29 & 38.77$\pm$5.79 & 28.42$\pm$1.58 & 31.21$\pm$2.39\\
    \bottomrule
    \multicolumn{5}{|c|}{Set 2}\\
    \bottomrule
    BERT & 66.30$\pm$4.52 & 32.42$\pm$2.08 & 28.41$\pm$2.95 & 29.09$\pm$3.08\\
    RoBERTa & 66.14$\pm$1.70 & 40.77$\pm$8.44 & 30.47$\pm$6.38 & 32.85$\pm$7.03\\
    RoBERTa-Hate & 68.10$\pm$1.57 & 39.93$\pm$4.37 & 32.18$\pm$4.57 & 33.81$\pm$4.63\\
    \bottomrule
    LLaMA & 64.79$\pm$1.97 & 54.62$\pm$3.75 & 51.77$\pm$3.83 & 52.15$\pm$3.65\\
    LLaMA w/ Def. & 64.70$\pm$2.37 & 53.11$\pm$11.04 & 53.98 $\pm$3.67 & 54.16$\pm$4.43\\
    GPT & 54.19$\pm$2.77 & \bf{55.61}$\pm$5.11 & 54.29$\pm$5.79 & 50.53$\pm$5.08\\
    GPT w/ Def. & 64.43$\pm$1.52 & 54.15$\pm$3.42 & \bf{60.02}$\pm$3.11 & \bf{55.98}$\pm$3.08\\
    \bottomrule
    \multicolumn{5}{|c|}{Set 3}\\
    \bottomrule
    BERT & 66.30$\pm$4.32 & 21.53$\pm$2.29 & 19.14$\pm$1.51 & 19.49$\pm$1.60\\
    RoBERTa & 65.84$\pm$2.24 & 30.52$\pm$8.89 & 23.01$\pm$6.86 & 23.60$\pm$6.55\\
    RoBERTa-Hate & 67.80$\pm$2.07 & 26.00$\pm$2.28 & 25.09$\pm$3.29 & 24.22$\pm$2.96\\
    \bottomrule
    \end{tabular}
    \caption{Accuracy, Precision, Recall and macro F1-Score (F1) for hostility type classification in flat (top) and 2-level hierarchical ways (bottom) ($\pm$ std. dev.). Best results are in bold.}
    \label{tab:multiclass_predictive_results}
\end{table}

We apply LLMs on Set 2, where better results are achieved. Among four LLM settings, LLaMA w/ Def. achieves the best performance with a macro F1 score of 72.91, followed by GPT w/ Def (70.69 F1). We notice that adding label definitions in the prompt leads to performance improvement (+1.80 F1 for LLaMA and +12.02 F1 for GPT). We argue that advanced LLMs do not show significant advantages on binary hostility identification as it is a simple and straightforward 2-class classification task.

\paragraph{Multi-class Hostility Classification} Table \ref{tab:multiclass_predictive_results} presents the results of all models on multi-class hostility type classification using three sets of data in flat (top) and 2-level hierarchical (bottom) ways. Among all PLMs, the best performing model is RoBERTa-Hate on Set 2 in the flat classification method with an F1 score of 43.31 (in bold). Similar to the binary hostility identification, models in Set 2 achieve the best predictive results compared with the same models trained on other sets (e.g., 32.04 F1 for RoBERTa in flat classification, 33.81 F1 for RoBERTa-Hate in hierarchical classification), followed by Set 1  (e.g., 20.79 F1 for RoBERTa in flat classification, 31.21 F1 for RoBERTa-Hate in hierarchical classification). The domain adaptation model, RoBERTa-Hate, outperforms the vanilla RoBERTa model with a larger difference compared to binary hostility identification (e.g., +4.17 F1 vs. +11.27 F1 on Set 2 in binary hostility identification and in multi-class hostility classification using the flat method). Additionally, BERT and RoBERTa exhibit comparable performance, with BERT sometimes outperforming RoBERTa and vice versa depending on different settings. 

Furthermore, the flat classification method outperforms the 2-level hierarchical classification method (e.g., 36.67 F1 in flat classification vs. 27.65 F1 in hierarchical for RoBERTa-Hate on Set 1). This may be explained by the fact that after training the first classifier, predictive errors introduce more noise in the hierarchical method. Also, the second classifier is trained on a smaller data set (non-hostile tweets are excluded) compared with that in the flat classification.

Similar to the hostility identification task, we only apply LLMs on Set 2. First of all, GPT w/ Def. in hierarchical classification outperforms all PLMs and LLMs, reaching a macro F1 score up to 55.98, which is 12.67 higher than the best performing PLM, RoBERTa-Hate. Secondly, in general, adding definitions of each hostility type boosts the performance. Moreover, prompts with definitions result in a larger improvement on the multi-class classification than the binary one (e.g., +5.85 F1 for LLaMA in flat classification, +5.45 F1 for GPT in hierarchical classification). Furthermore, LLMs in flat and hierarchical settings achieve comparable results. This might be explained by the fact that we use zero-shot classification, where they do not rely on our training data. Future research may extend it to few-shot learning. 

\section{Conclusion}
This work focuses on investigating online hostility towards UK politicians. We develop an English dataset of 3,320 tweets, which are manually annotated with hostility as well as their targeted identity characteristics: religion, gender, and race. Also, we conduct extensive linguistic and topical analyses to provide deeper insights into the specific content of these hostile interactions. By constructing and analysing such a dataset, we identify key patterns, such as the prevalence of race-based hostility, especially regarding immigration issues in the UK. Also, our findings suggest that there is a general lack of trust in MPs in the UK. Furthermore, we evaluate various PLMs and LLMs on binary hostility identification and multi-class targeted identity type classification in flat and hierarchical ways. This study not only offers valuable data but also lays the groundwork for future research aimed at understanding and mitigating the impact of online hostility in political contexts specific to the UK. 

\section{Acknowledgments}
The study was conducted as part of the ``Responsible AI for Inclusive, Democratic Societies: A cross-disciplinary approach to detecting and countering abusive language online'' project [grant number R/163157-11-1]. 

\bibliography{aaai22}

\begin{thebibliography}{56}
\providecommand{\natexlab}[1]{#1}

\bibitem[{Agarwal et~al.(2021)Agarwal, Hawkins, Amaxopoulou, Dempsey, Sastry, and Wood}]{agarwal2021hate}
Agarwal, P.; Hawkins, O.; Amaxopoulou, M.; Dempsey, N.; Sastry, N.; and Wood, E. 2021.
\newblock Hate speech in political discourse: A case study of UK MPs on Twitter.
\newblock In \emph{Proceedings of the 32nd ACM conference on hypertext and social media}, 5--16.

\bibitem[{Agarwal, Sastry, and Wood(2019)}]{agarwal2019tweeting}
Agarwal, P.; Sastry, N.; and Wood, E. 2019.
\newblock Tweeting mps: Digital engagement between citizens and members of parliament in the uk.
\newblock In \emph{Proceedings of the International AAAI Conference on Web and Social Media}, volume~13, 26--37.

\bibitem[{Amarasingam, Umar, and Desai(2022)}]{amarasingam2022fight}
Amarasingam, A.; Umar, S.; and Desai, S. 2022.
\newblock “Fight, die, and if required kill”: Hindu nationalism, misinformation, and Islamophobia in India.
\newblock \emph{Religions}, 13(5): 380.

\bibitem[{Artstein and Poesio(2008)}]{artstein2008inter}
Artstein, R.; and Poesio, M. 2008.
\newblock {Inter-Coder Agreement for Computational Linguistics}.
\newblock \emph{Computational linguistics}, 34(4): 555--596.

\bibitem[{Bakir, Farrell, and Bontcheva(2024)}]{bakir2024abuse}
Bakir, M.~E.; Farrell, T.; and Bontcheva, K. 2024.
\newblock Abuse in the time of COVID-19: the effects of Brexit, gender and partisanship.
\newblock \emph{Online Information Review}.

\bibitem[{Basile et~al.(2019)Basile, Bosco, Fersini, Nozza, Patti, Pardo, Rosso, and Sanguinetti}]{basile2019semeval}
Basile, V.; Bosco, C.; Fersini, E.; Nozza, D.; Patti, V.; Pardo, F. M.~R.; Rosso, P.; and Sanguinetti, M. 2019.
\newblock Semeval-2019 task 5: Multilingual detection of hate speech against immigrants and women in twitter.
\newblock In \emph{Proceedings of the 13th international workshop on semantic evaluation}, 54--63.

\bibitem[{Bird, Klein, and Loper(2009)}]{bird2009natural}
Bird, S.; Klein, E.; and Loper, E. 2009.
\newblock \emph{Natural language processing with Python: analyzing text with the natural language toolkit}.
\newblock " O'Reilly Media, Inc.".

\bibitem[{Boyd et~al.(2022)Boyd, Ashokkumar, Seraj, and Pennebaker}]{boyd2022development}
Boyd, R.~L.; Ashokkumar, A.; Seraj, S.; and Pennebaker, J.~W. 2022.
\newblock The development and psychometric properties of LIWC-22.
\newblock \emph{Austin, TX: University of Texas at Austin}, 1--47.

\bibitem[{Carson et~al.(2024)Carson, Mikolajczak, Ruppanner, and Foley}]{carson2024online}
Carson, A.; Mikolajczak, G.; Ruppanner, L.; and Foley, E. 2024.
\newblock From online trolls to ‘Slut Shaming’: Understanding the role of incivility and gender abuse in local government.
\newblock \emph{Local Government Studies}, 50(2): 427--450.

\bibitem[{Collignon and R{\"u}dig(2021)}]{collignon2021increasing}
Collignon, S.; and R{\"u}dig, W. 2021.
\newblock Increasing the cost of female representation? The gendered effects of harassment, abuse and intimidation towards Parliamentary candidates in the UK.
\newblock \emph{Journal of elections, public opinion and parties}, 31(4): 429--449.

\bibitem[{Enock et~al.(2023)Enock, Johansson, Bright, and Margetts}]{enock2023tracking}
Enock, F.; Johansson, P.; Bright, J.; and Margetts, H.~Z. 2023.
\newblock Tracking experiences of online harms and attitudes towards online safety interventions: Findings from a large-scale, nationally representative survey of the british public.
\newblock \emph{Nationally Representative Survey of the British Public (March 21, 2023)}.

\bibitem[{Esposito and Breeze(2022)}]{esposito2022gender}
Esposito, E.; and Breeze, R. 2022.
\newblock Gender and politics in a digitalised world: Investigating online hostility against UK female MPs.
\newblock \emph{Discourse \& Society}, 33(3): 303--323.

\bibitem[{Esposito and Zollo(2021)}]{esposito2021dare}
Esposito, E.; and Zollo, S.~A. 2021.
\newblock “How dare you call her a pig, I know several pigs who would be upset if they knew” A multimodal critical discursive approach to online misogyny against UK MPs on YouTube.
\newblock \emph{Journal of language aggression and conflict}, 9(1): 47--75.

\bibitem[{Farrell, Bakir, and Bontcheva(2021)}]{farrell2021mp}
Farrell, T.; Bakir, M.; and Bontcheva, K. 2021.
\newblock Mp twitter engagement and abuse post-first covid-19 lockdown in the uk: White paper.
\newblock \emph{arXiv preprint arXiv:2103.02917}.

\bibitem[{Fleiss(1971)}]{fleiss1971measuring}
Fleiss, J.~L. 1971.
\newblock Measuring nominal scale agreement among many raters.
\newblock \emph{Psychological bulletin}, 76(5): 378.

\bibitem[{{FORCE11}(2020)}]{fair}
{FORCE11}. 2020.
\newblock The FAIR Data principles.
\newblock \url{https://force11.org/info/the-fair-data-principles/}.

\bibitem[{Fortuna, Soler, and Wanner(2020)}]{fortuna2020toxic}
Fortuna, P.; Soler, J.; and Wanner, L. 2020.
\newblock Toxic, hateful, offensive or abusive? what are we really classifying? an empirical analysis of hate speech datasets.
\newblock In \emph{Proceedings of the Twelfth Language Resources and Evaluation Conference}, 6786--6794.

\bibitem[{Fuchs and Sch{\"a}fer(2021)}]{fuchs2021normalizing}
Fuchs, T.; and Sch{\"a}fer, F. 2021.
\newblock Normalizing misogyny: hate speech and verbal abuse of female politicians on Japanese Twitter.
\newblock In \emph{Japan forum}, volume~33, 553--579. Taylor \& Francis.

\bibitem[{Goodman and Locke(2024)}]{goodman2024supporting}
Goodman, S.; and Locke, A. 2024.
\newblock Supporting and challenging hate in an online discussion of a controversial refugee policy.
\newblock \emph{Discourse Studies}, 14614456231225448.

\bibitem[{Gorrell et~al.(2019)Gorrell, Bakir, Greenwood, Roberts, and Bontcheva}]{gorrell2019race}
Gorrell, G.; Bakir, M.~E.; Greenwood, M.~A.; Roberts, I.; and Bontcheva, K. 2019.
\newblock Race and Religion in Online Abuse towards UK Politicians: Working Paper.
\newblock \emph{arXiv preprint ArXiv:1910.00920 [Cs]}.

\bibitem[{Gorrell et~al.(2020)Gorrell, Bakir, Roberts, Greenwood, and Bontcheva}]{gorrell2020politicians}
Gorrell, G.; Bakir, M.~E.; Roberts, I.; Greenwood, M.~A.; and Bontcheva, K. 2020.
\newblock Which politicians receive abuse? Four factors illuminated in the UK general election 2019.
\newblock \emph{EPJ Data Science}, 9(1): 18.

\bibitem[{Gorrell et~al.(2018)Gorrell, Greenwood, Roberts, Maynard, and Bontcheva}]{gorrell2018twits}
Gorrell, G.; Greenwood, M.; Roberts, I.; Maynard, D.; and Bontcheva, K. 2018.
\newblock Twits, twats and twaddle: Trends in online abuse towards uk politicians.
\newblock In \emph{Proceedings of the International AAAI Conference on Web and Social Media}, volume~12.

\bibitem[{Grimminger and Klinger(2021)}]{grimminger2021hate}
Grimminger, L.; and Klinger, R. 2021.
\newblock Hate Towards the Political Opponent: A {T}witter Corpus Study of the 2020 {US} Elections on the Basis of Offensive Speech and Stance Detection.
\newblock In \emph{Proceedings of the Eleventh Workshop on Computational Approaches to Subjectivity, Sentiment and Social Media Analysis}, 171--180. Online: Association for Computational Linguistics.

\bibitem[{Grootendorst(2022)}]{grootendorst2022bertopic}
Grootendorst, M. 2022.
\newblock BERTopic: Neural topic modeling with a class-based TF-IDF procedure.
\newblock \emph{arXiv preprint arXiv:2203.05794}.

\bibitem[{Gross et~al.(2023)Gross, Baltz, Suttmann-Lea, Merivaki, and Stewart~III}]{gross2023online}
Gross, J.; Baltz, S.; Suttmann-Lea, M.; Merivaki, L.; and Stewart~III, C. 2023.
\newblock Online Hostility Towards Local Election Officials Surged in 2020.
\newblock \emph{Available at SSRN 4351996}.

\bibitem[{Guellil et~al.(2020)Guellil, Adeel, Azouaou, Chennoufi, Maafi, and Hamitouche}]{guellil2020detecting}
Guellil, I.; Adeel, A.; Azouaou, F.; Chennoufi, S.; Maafi, H.; and Hamitouche, T. 2020.
\newblock Detecting hate speech against politicians in Arabic community on social media.
\newblock \emph{International Journal of Web Information Systems}, 16(3): 295--313.

\bibitem[{H{\aa}kansson(2024)}]{haakansson2024explaining}
H{\aa}kansson, S. 2024.
\newblock Explaining citizen hostility against women political leaders: A survey experiment in the United States and Sweden.
\newblock \emph{Politics \& Gender}, 20(1): 1--28.

\bibitem[{Hartvigsen et~al.(2022)Hartvigsen, Gabriel, Palangi, Sap, Ray, and Kamar}]{hartvigsen2022toxigen}
Hartvigsen, T.; Gabriel, S.; Palangi, H.; Sap, M.; Ray, D.; and Kamar, E. 2022.
\newblock ToxiGen: A Large-Scale Machine-Generated Dataset for Adversarial and Implicit Hate Speech Detection.
\newblock In \emph{Proceedings of the 60th Annual Meeting of the Association for Computational Linguistics (Volume 1: Long Papers)}, 3309--3326.

\bibitem[{Hua, Naaman, and Ristenpart(2020)}]{hua2020characterizing}
Hua, Y.; Naaman, M.; and Ristenpart, T. 2020.
\newblock Characterizing twitter users who engage in adversarial interactions against political candidates.
\newblock In \emph{Proceedings of the 2020 CHI conference on human factors in computing systems}, 1--13.

\bibitem[{Jafri et~al.(2023)Jafri, Siddiqui, Thapa, Rauniyar, Naseem, and Razzak}]{jafri2023uncovering}
Jafri, F.~A.; Siddiqui, M.~A.; Thapa, S.; Rauniyar, K.; Naseem, U.; and Razzak, I. 2023.
\newblock Uncovering Political Hate Speech During Indian Election Campaign: A New Low-Resource Dataset and Baselines.
\newblock \emph{arXiv e-prints}.

\bibitem[{Jahan and Oussalah(2023)}]{jahan2023systematic}
Jahan, M.~S.; and Oussalah, M. 2023.
\newblock A systematic review of Hate Speech automatic detection using Natural Language Processing.
\newblock \emph{Neurocomputing}, 126232.

\bibitem[{Jin et~al.(2023)Jin, Mu, Maynard, and Bontcheva}]{jin2023examining}
Jin, M.; Mu, Y.; Maynard, D.; and Bontcheva, K. 2023.
\newblock Examining temporal bias in abusive language detection.
\newblock \emph{arXiv preprint arXiv:2309.14146}.

\bibitem[{Kenton and Toutanova(2019)}]{devlin2018bert}
Kenton, J. D. M.-W.~C.; and Toutanova, L.~K. 2019.
\newblock Bert: Pre-training of deep bidirectional transformers for language understanding.
\newblock In \emph{Proceedings of naacL-HLT}, volume~1, 2.

\bibitem[{Kuperberg(2018)}]{kuperberg2018intersectional}
Kuperberg, R. 2018.
\newblock Intersectional violence against women in politics.
\newblock \emph{Politics \& Gender}, 14(4): 685--690.

\bibitem[{Kuperberg(2021)}]{kuperberg2021incongruous}
Kuperberg, R. 2021.
\newblock Incongruous and illegitimate: Antisemitic and Islamophobic semiotic violence against women in politics in the United Kingdom.
\newblock \emph{Journal of Language Aggression and Conflict}, 9(1): 100--126.

\bibitem[{Kwarteng et~al.(2022)Kwarteng, Perfumi, Farrell, Third, and Fernandez}]{kwarteng2022misogynoir}
Kwarteng, J.; Perfumi, S.~C.; Farrell, T.; Third, A.; and Fernandez, M. 2022.
\newblock Misogynoir: challenges in detecting intersectional hate.
\newblock \emph{Social Network Analysis and Mining}, 12(1): 166.

\bibitem[{Lavalley and Johnson(2022)}]{lavalley2022occupation}
Lavalley, R.; and Johnson, K.~R. 2022.
\newblock Occupation, injustice, and anti-Black racism in the United States of America.
\newblock \emph{Journal of Occupational Science}, 29(4): 487--499.

\bibitem[{Liu et~al.(2019)Liu, Ott, Goyal, Du, Joshi, Chen, Levy, Lewis, Zettlemoyer, and Stoyanov}]{liu2019roberta}
Liu, Y.; Ott, M.; Goyal, N.; Du, J.; Joshi, M.; Chen, D.; Levy, O.; Lewis, M.; Zettlemoyer, L.; and Stoyanov, V. 2019.
\newblock RoBERTa: A Robustly Optimized BERT Pretraining Approach.
\newblock \emph{arXiv preprint arXiv:1907.11692}.

\bibitem[{MacAvaney et~al.(2019)MacAvaney, Yao, Yang, Russell, Goharian, and Frieder}]{macavaney2019hate}
MacAvaney, S.; Yao, H.-R.; Yang, E.; Russell, K.; Goharian, N.; and Frieder, O. 2019.
\newblock Hate speech detection: Challenges and solutions.
\newblock \emph{PloS one}, 14(8): e0221152.

\bibitem[{Mansur, Omar, and Tiun(2023)}]{mansur2023twitter}
Mansur, Z.; Omar, N.; and Tiun, S. 2023.
\newblock Twitter hate speech detection: a systematic review of methods, taxonomy analysis, challenges, and opportunities.
\newblock \emph{IEEE Access}, 11: 16226--16249.

\bibitem[{Mathew et~al.(2021)Mathew, Saha, Yimam, Biemann, Goyal, and Mukherjee}]{mathew2021hatexplain}
Mathew, B.; Saha, P.; Yimam, S.~M.; Biemann, C.; Goyal, P.; and Mukherjee, A. 2021.
\newblock Hatexplain: A benchmark dataset for explainable hate speech detection.
\newblock In \emph{Proceedings of the AAAI conference on artificial intelligence}, volume~35, 14867--14875.

\bibitem[{Mollas et~al.(2022)Mollas, Chrysopoulou, Karlos, and Tsoumakas}]{mollas2022ethos}
Mollas, I.; Chrysopoulou, Z.; Karlos, S.; and Tsoumakas, G. 2022.
\newblock ETHOS: a multi-label hate speech detection dataset.
\newblock \emph{Complex \& Intelligent Systems}, 8(6): 4663--4678.

\bibitem[{Pavlopoulos et~al.(2020)Pavlopoulos, Sorensen, Dixon, Thain, and Androutsopoulos}]{pavlopoulos2020toxicity}
Pavlopoulos, J.; Sorensen, J.; Dixon, L.; Thain, N.; and Androutsopoulos, I. 2020.
\newblock Toxicity Detection: Does Context Really Matter?
\newblock In \emph{Proceedings of the 58th Annual Meeting of the Association for Computational Linguistics}, 4296--4305.

\bibitem[{Rieger et~al.(2021)Rieger, K{\"u}mpel, Wich, Kiening, and Groh}]{rieger2021assessing}
Rieger, D.; K{\"u}mpel, A.~S.; Wich, M.; Kiening, T.; and Groh, G. 2021.
\newblock Assessing the extent and types of hate speech in fringe communities: A case study of alt-right communities on 8chan, 4chan, and Reddit.
\newblock \emph{Social Media+ Society}, 7(4): 20563051211052906.

\bibitem[{Rosa et~al.(2019)Rosa, Pereira, Ribeiro, Ferreira, Carvalho, Oliveira, Coheur, Paulino, Sim{\~a}o, and Trancoso}]{rosa2019automatic}
Rosa, H.; Pereira, N.; Ribeiro, R.; Ferreira, P.~C.; Carvalho, J.~P.; Oliveira, S.; Coheur, L.; Paulino, P.; Sim{\~a}o, A.~V.; and Trancoso, I. 2019.
\newblock Automatic cyberbullying detection: A systematic review.
\newblock \emph{Computers in Human Behavior}, 93: 333--345.

\bibitem[{R{\"o}ttger et~al.(2022)R{\"o}ttger, Vidgen, Hovy, and Pierrehumbert}]{rottger2021two}
R{\"o}ttger, P.; Vidgen, B.; Hovy, D.; and Pierrehumbert, J. 2022.
\newblock Two Contrasting Data Annotation Paradigms for Subjective NLP Tasks.
\newblock In \emph{Proceedings of the 2022 Conference of the North American Chapter of the Association for Computational Linguistics: Human Language Technologies}, 175--190.

\bibitem[{Scott(2019)}]{scott2019women}
Scott, J. 2019.
\newblock Women MPs say abuse forcing them from politics.
\newblock \emph{BBC News}.

\bibitem[{Solovev and Pr{\"o}llochs(2022)}]{solovev2022hate}
Solovev, K.; and Pr{\"o}llochs, N. 2022.
\newblock Hate speech in the political discourse on social media: Disparities across parties, gender, and ethnicity.
\newblock In \emph{Proceedings of the ACM Web Conference 2022}, 3656--3661.

\bibitem[{Southern and Harmer(2021)}]{southern2021twitter}
Southern, R.; and Harmer, E. 2021.
\newblock Twitter, incivility and “everyday” gendered othering: An analysis of tweets sent to UK members of parliament.
\newblock \emph{Social science computer review}, 39(2): 259--275.

\bibitem[{Vidgen and Yasseri(2020)}]{vidgen2020detecting}
Vidgen, B.; and Yasseri, T. 2020.
\newblock Detecting weak and strong Islamophobic hate speech on social media.
\newblock \emph{Journal of Information Technology \& Politics}, 17(1): 66--78.

\bibitem[{Walther(2022)}]{walther2022social}
Walther, J.~B. 2022.
\newblock Social media and online hate.
\newblock \emph{Current Opinion in Psychology}, 45: 101298.

\bibitem[{Wang, Day, and Wu(2022)}]{wang2022political}
Wang, C.-C.; Day, M.-Y.; and Wu, C.-L. 2022.
\newblock Political hate speech detection and lexicon building: A study in taiwan.
\newblock \emph{IEEE Access}, 10: 44337--44346.

\bibitem[{Ward and McLoughlin(2020)}]{ward2020turds}
Ward, S.; and McLoughlin, L. 2020.
\newblock Turds, traitors and tossers: the abuse of UK MPs via Twitter.
\newblock \emph{The Journal of Legislative Studies}, 26(1): 47--73.

\bibitem[{Waseem et~al.(2017)Waseem, Davidson, Warmsley, and Weber}]{waseem2017understanding}
Waseem, Z.; Davidson, T.; Warmsley, D.; and Weber, I. 2017.
\newblock Understanding Abuse: A Typology of Abusive Language Detection Subtasks.
\newblock In \emph{Proceedings of the First Workshop on Abusive Language Online}, 78--84.

\bibitem[{Wilby et~al.(2023)Wilby, Karmakharm, Roberts, Song, and Bontcheva}]{wilby2023gate}
Wilby, D.; Karmakharm, T.; Roberts, I.; Song, X.; and Bontcheva, K. 2023.
\newblock GATE Teamware 2: An open-source tool for collaborative document classification annotation.
\newblock In \emph{Proceedings of the 17th Conference of the European Chapter of the Association for Computational Linguistics: System Demonstrations}, 145--151.

\bibitem[{Zampieri et~al.(2019)Zampieri, Malmasi, Nakov, Rosenthal, Farra, and Kumar}]{zampieri2019semeval}
Zampieri, M.; Malmasi, S.; Nakov, P.; Rosenthal, S.; Farra, N.; and Kumar, R. 2019.
\newblock SemEval-2019 Task 6: Identifying and Categorizing Offensive Language in Social Media (OffensEval).
\newblock In \emph{Proceedings of the 13th International Workshop on Semantic Evaluation}, 75--86.

\end{thebibliography}

\appendix
\section{Dataset Availability}
Our dataset is publicly available in accordance with the FAIR principles \cite{fair}:
\begin{itemize}
    \item \textbf{Findable:} Our dataset will be published in the Zenodo dataset-sharing service with a unique digital object identifier. For now it can be found at \url{https://anonymous.4open.science/r/ohtukmp-21D8}.
    \item \textbf{Accessible:} Original tweets can be retrieved using their tweet IDs via the standard Twitter API\footnote{https://developer.twitter.com/en/docs/twitter-api/tweets/lookup/api-reference/get-tweets-id}
    \item \textbf{Interoperable:} The readme file in the repository explains the dataset structure and the description of each column in the CSV data file. CSV datasets are easily imported and processed by most widely used data processing tools.
    \item \textbf{Re-usable:} Our dataset can be re-used by anyone who has Twitter developer account.
\end{itemize}
\section{Model Parameters}
For BERT, we use base uncased model, and for RoBERTa, we use base model. The maximum sequence length is set to 256 tokens, and the batch size is set to 32. We run all models using a 5-fold cross-validation method where 4-fold data is used for training and 1-fold data is used for testing. We split 4-fold data into training and validation sets with a ratio of 9:1. We use Cross Entropy Loss as the training loss function with the AdamW optimizer. For flat classification, we train using a learning rate of lr = 3e-6, while hierarchical classification is trained with lr = 5e-5. During training, we choose the model with the smallest validation loss over 15 epochs. All models are trained on an NVIDIA A100 GPU. For all experiments of LLMs, the temperature is set to 0.1
\section{Confidence Scores}
Table \ref{tab:confidence} shows the meaning and explanation of different levels of confidence scores (1-5) used in the annotation task.

\begin{table}[t!]

    \small
\begin{tabular}{|c|l|p{0.5\linewidth}|}
    \toprule
    \bf{Score} & \bf{Meaning} & \bf{Explanation}\\
    \midrule
        5 & Extremely confident & I'm certain without a doubt.  \\
        4 & Fairly confident & I’m confident, but there might be a small chance other annotators may label it as a different category. \\
        3 & Pretty confident & I’m pretty sure, but there might be a high chance other annotators may label it as a different category. \\
        2 & Not confident & I’m not sure; it could belong to this or another category. \\
        1 & Very low confidence & I’m really unsure; it might belong to another category instead.\\
        \bottomrule
    \end{tabular}
    \caption{Confidence Scores used in the annotation task.}
    \label{tab:confidence}
\end{table}
\end{document}